\documentclass[conference]{IEEEtran}
\IEEEoverridecommandlockouts
\usepackage{cite}
\usepackage{amsmath,amssymb,amsfonts}
\usepackage{algorithmic}
\usepackage{graphicx}
\usepackage{textcomp}
\usepackage{xcolor}
\usepackage{subfigure}
\def\BibTeX{{\rm B\kern-.05em{\sc i\kern-.025em b}\kern-.08em
    T\kern-.1667em\lower.7ex\hbox{E}\kern-.125emX}}
\begin{document}

\title{A Cloud-Edge-Terminal Collaborative System for Temperature Measurement in COVID-19 Prevention}

\author{\IEEEauthorblockN{
		Zheyi Ma\IEEEauthorrefmark{1},
		Hao Li\IEEEauthorrefmark{1},
		Wen Fang\IEEEauthorrefmark{1},
		Qingwen Liu\IEEEauthorrefmark{1},
		Bin Zhou\IEEEauthorrefmark{2} and
		Zhiyong Bu\IEEEauthorrefmark{2}
	}
	
	\IEEEauthorblockA{\IEEEauthorrefmark{1}Dept. of Computer Science and Technology, Tongji University, Shanghai, China}
	
	\IEEEauthorblockA{\IEEEauthorrefmark{2}Key Laboratory of Wireless Sensor Network and Communications,\\	
		Shanghai Institute of Microsystem and Information Technology, Chinese Academy of Sciences, Shanghai, China}
	
	\IEEEauthorblockA{Email: \{zheyima, lihao1101, wen.fang, qliu\}@tongji.edu.cn, \{bin.zhou, zhiyong.bu\}@mail.sim.ac.cn}

	\thanks{The work was supported by the National Key Research and Development Project under Grant 2020YFB2103900 and Grant 2020YFB2103902. It was also supported by the National Natural Science Foundation of China under Grant 61771344 and Grant 62071334.}	
}

\maketitle

\begin{abstract}
To prevent the spread of coronavirus disease 2019 (COVID-19), preliminary temperature measurement and mask detection in public areas are conducted. However, the existing temperature measurement methods face the problems of safety and deployment. In this paper, to realize safe and accurate temperature measurement even when a person's face is partially obscured, we propose a cloud-edge-terminal collaborative system with a lightweight infrared temperature measurement model. A binocular camera with an RGB lens and a thermal lens is utilized to simultaneously capture image pairs. Then, a mobile detection model based on a multi-task cascaded convolutional network (MTCNN) is proposed to realize face alignment and mask detection on the RGB images. For accurate temperature measurement, we transform the facial landmarks on the RGB images to the thermal images by an affine transformation and select a more accurate temperature measurement area on the forehead. The collected information is uploaded to the cloud in real time for COVID-19 prevention. Experiments show that the detection model is only 6.1M and the average detection speed is 257ms. At a distance of 1m, the error of indoor temperature measurement is about 3\%. That is, the proposed system can realize real-time temperature measurement in public areas. 
\end{abstract}

\begin{IEEEkeywords}
COVID-19 Prevention, Mobile Temperature Measurement, Cloud-Edge-Terminal Collaborative System
\end{IEEEkeywords}

\section{Introduction}
\label{sec:Introduction}
In December 2019, a novel coronavirus named coronavirus disease 2019 (COVID-19) was identified in Wuhan, China, and spread quickly around the world. A year later in December 2020, more than 70 million people were infected with the virus, causing more than 1.6 million deaths~\cite{COVID-19dashboard}. The research\cite{COVID-19features} shows that  the clinical features of COVID-19 pneumonia are similar to other pneumonia, but liver function damage is more frequent in COVID-19 than in  non-COVID-19 patients. That is, COVID-19 is a more infectious and dangerous disease.

Fever is an initial symptom of COVID-19\cite{COVID-19}. Therefore, temperature measurement is an effective way to find patients. Nowadays, pedestrians are forced to measure their body temperature before entering public areas, such as subway stations and train stations. 
There are two main methods of temperature measurement:

\begin{itemize}	
	\item Handheld infrared thermometer for short-distance temperature measurement.
	
	\item Thermal camera for remote temperature measurement.
\end{itemize}

The infrared thermometer is cheap and easy to operate, but it requires close contact between the inspector and pedestrians. It is unsafe when pedestrians carry the virus. To realize safe temperature measurement, thermal cameras which can measure temperature from a long distance have been developed. However, the current design of thermal cameras usually requires the purchase of other hardware, such as displays and computers, to support the operation of the system. In addition, body temperature information cannot be used by the system for COVID-19 prevention effectively and timely. Finally, to prevent the spread of the virus through respiratory droplets, people always wear masks in public areas. Thus, mask detection in public areas needs to be taken into account. 

\begin{figure}[htbp]
	\centering
	\includegraphics[width=2.6in]{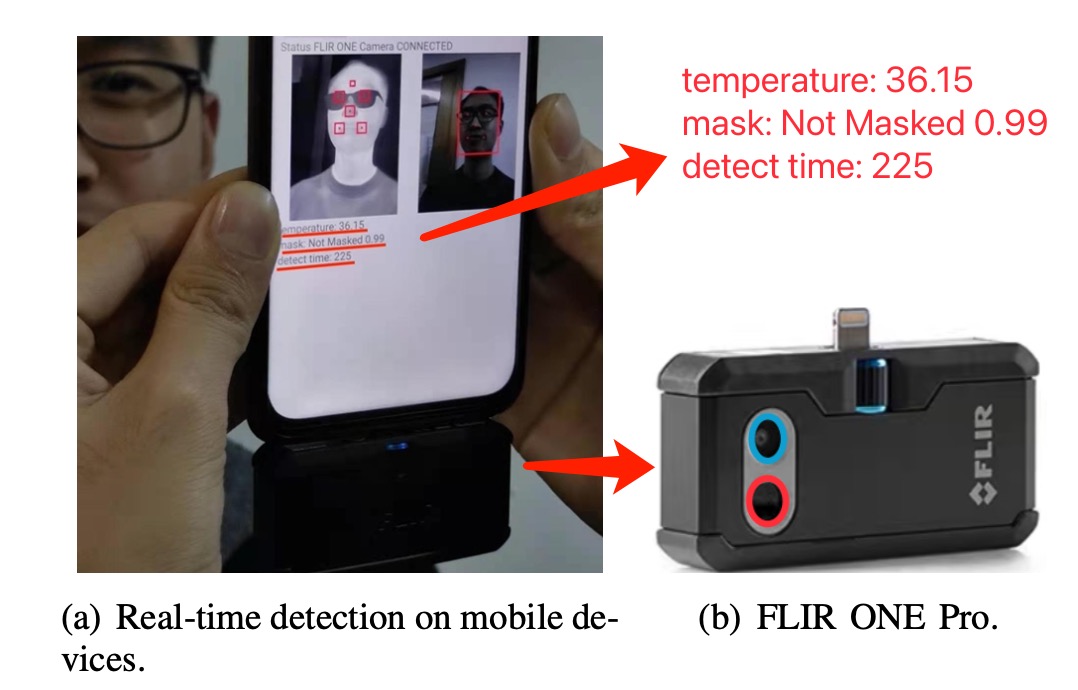}
	\caption{Real-time detection on the mobile device.}
	\label{fig:scenario}
\end{figure}

A skin temperature extraction method, proposed by Aryal in~\cite{aryal2019skin}, uses the binocular camera with an RGB lens and a thermal lens to detect people's skin temperature. However, the occlusion problem and the method capable of accurately positioning temperature measurement areas were not analyzed in the paper. In this paper, to accurately measure body temperature even when a person's face is partially obscured, we propose a cloud-edge-terminal collaborative system for remote temperature measurement and develop a detection model which can select a more accurate temperature measurement area on the forehead. As shown in Fig.~\ref{fig:scenario}, the detection model can achieve real-time detection on mobile devices.
From the RGB and thermal images, we can get the person's temperature and whether he is wearing a mask. Finally, the cloud-edge-terminal collaborative system uploads the location and body temperature information to the cloud.

The contributions of this paper include:
\begin{itemize}	
	\item We propose a cloud-edge-terminal collaborative system, which has a lightweight face alignment model with a mask detection branch and is easy to deploy in mobile devices. The system can upload location and body temperature information to the cloud in real time.
	
	\item An accurate affine transformation matrix is calculated to align the image pairs taken by the binocular camera.
	
	\item Experiments show that our model can achieve real-time temperature measurement on mobile devices. The detection model is only 6.1M and the average detection speed is 257ms. The error of indoor temperature measurement is about 3\% at a distance of 1m.
	
\end{itemize}

In the rest of this paper, we show the design of the cloud-edge-terminal collaborative system in section~\ref{sec:collaborativeSystem}.
In section~\ref{sec:measurementMethod}, we present the infrared temperature measurement method. 
Then, a face alignment model with a mask detection branch is proposed and described in section~\ref{sec:MTCNNWithMaskDetection}.
In addition, we depict our experiments in section~\ref{sec:ExperimentsAndAnalysis}. 
Finally, we make a conclusion in section~\ref{sec:Conclusion}.

\section{Cloud-Edge-Terminal Collaborative System}
\label{sec:collaborativeSystem}
In this section, a cloud-edge-terminal collaborative system is proposed. With the help of this system, the timeliness of data transmission and the stability of the temperature measurement system can be ensured.

\begin{figure}[htbp]
	\centering
	\includegraphics[width=1.65in]{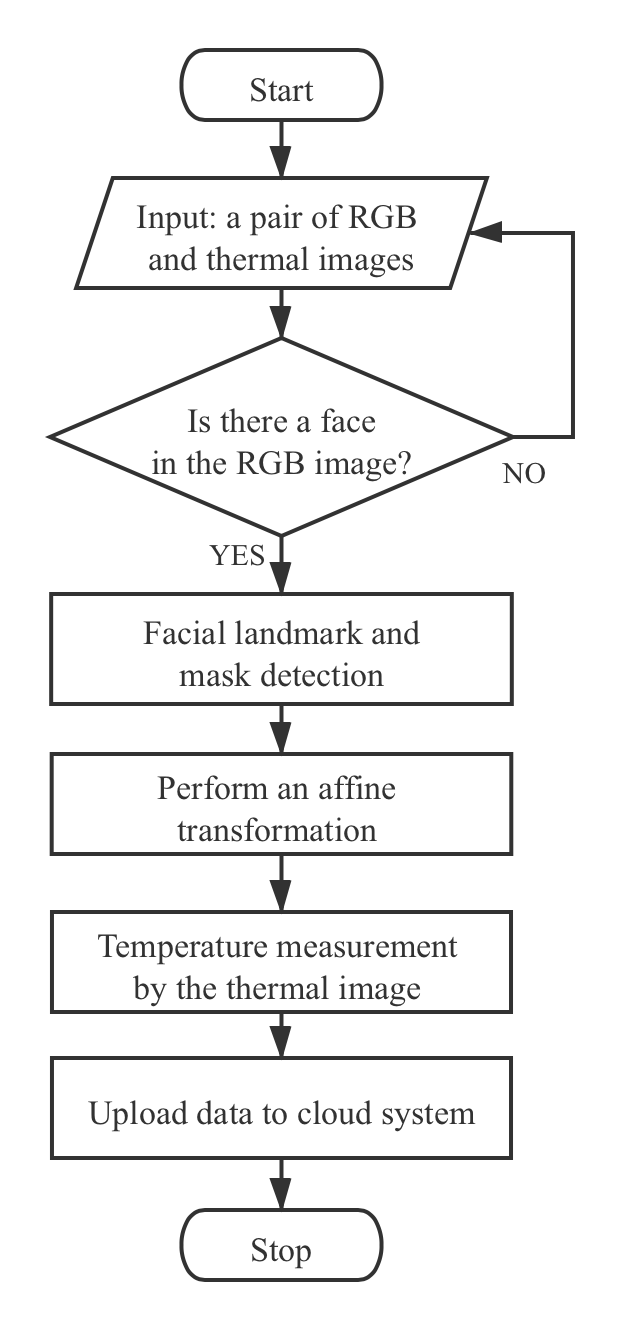}
	\caption{Flow chart of temperature measurement.}
	\label{fig:flowChart4}
\end{figure}

The work process of the cloud-edge-terminal collaborative system is shown in Fig.~\ref{fig:flowChart4}. At first, the input of the system is a pair of RGB and thermal images captured at the same time. Secondly, if faces exist in the RGB image, the facial landmark detection model can locate the landmarks. Otherwise, the camera retakes the valid images. Next, an affine transformation is performed and facial landmarks on the RGB image can be transformed into facial landmarks on the thermal image. Then, these facial landmarks are used for temperature measurement in the thermal image. Finally, the location and temperature information are uploaded to the cloud.

\subsection{Image Capturing}

As shown in Fig.~\ref{fig:scenario}(b), FLIR ONE Pro is a professional thermal camera, which is a binocular camera with an RGB lens marked by a blue circle and a thermal lens with a red circle~\cite{flironepro}. Due to the demand for mobile device deployment, we choose FLIR ONE Pro to capture image pairs. In addition, FLIR ONE Pro has an accuracy of $\pm3^{\circ}$C or $\pm5\%$ for temperature measure when the device is between 15$^{\circ}$C and 35$^{\circ}$C and the scene is between 5$^{\circ}$C and 120$^{\circ}$C. That is, the recommended working environment temperature meets the needs of daily body temperature detection.

Moreover, the thermal camera can take a pair of RGB and thermal images at the same time. The RGB image resolution is $1440 \times 1080$ pixels and the thermal image resolution is $640 \times 480$ pixels. We can get facial landmarks from the RGB images and get body temperature from the thermal images.

\subsection{Cloud-Edge-Terminal Collaborative System}
\begin{figure}[htbp]
	\centering
	\includegraphics[width=2.6in]{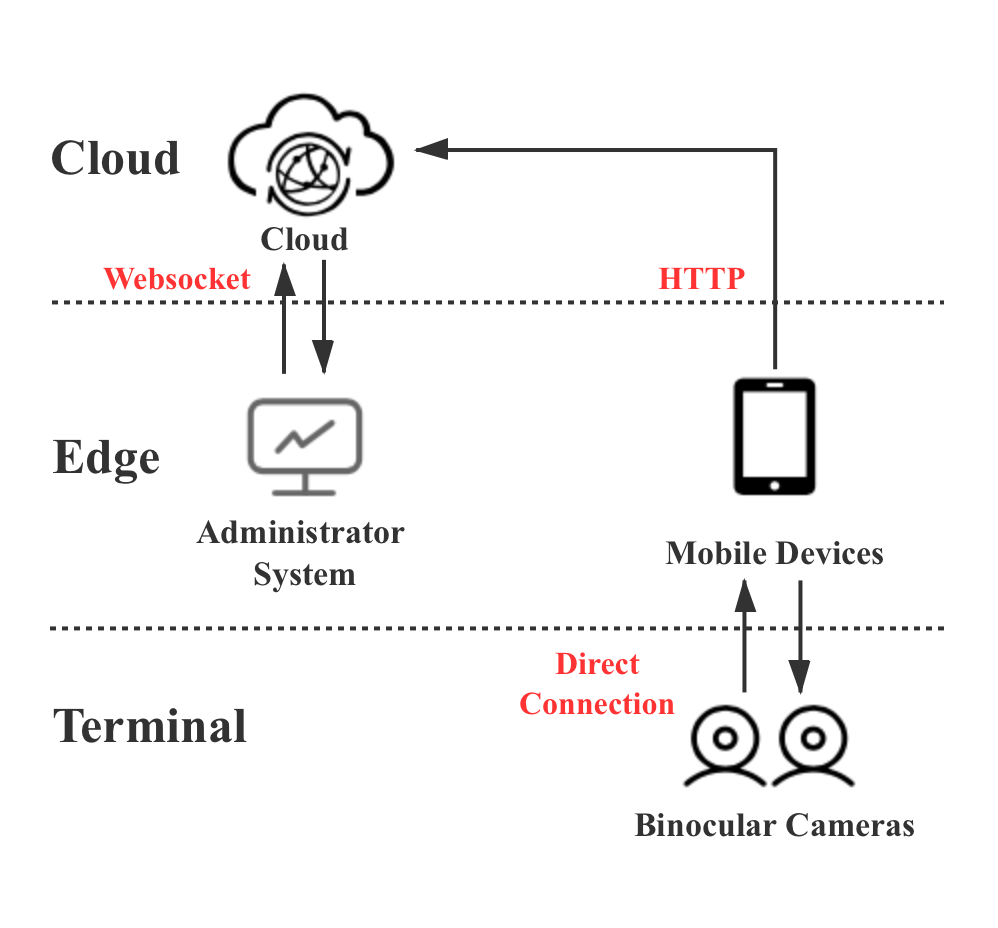}
	\caption{Cloud-edge-terminal collaborative system design.}
	\label{fig:systemDesign}
\end{figure}

After aligning the image pairs, we can select an accurate area for temperature. However, a small amount of data cannot be mined for additional information, we need to collect data to analyze COVID-19. To improve the efficiency of information transmission, a cloud-edge-terminal collaborative system shown in Fig.~\ref{fig:systemDesign} is necessary. 

\subsubsection{Cloud and  Mobile Devices Communication}
The binocular camera is directly connected to the mobile device through Type-C port, and the computing power of mobile devices is utilized to process the captured images. Once a pedestrian is detected, we record the position of the mobile device and the temperature of the pedestrian. The position and temperature of pedestrians can be uploaded to the cloud in real time by hypertext transfer protocol (HTTP)~\cite{tcp}, if the pedestrian’s body temperature is abnormal.

\subsubsection{Cloud and Administrator System Communication}
HTTP is a fundamental web protocol, and information can be sent to the cloud in one-way via HTTP. However, the cloud cannot actively send information to the administrator.
The WebSocket Protocol, proposed in~\cite{websocket}, enables two-way communication between a client and a host. 
We utilize WebSocket to realize full-duplex communication between the cloud and the administrator system. When the temperature exceeds the threshold, the cloud will immediately send an alert signal to the administrator system.


\section{Infrared Temperature Measurement Method}
\label{sec:measurementMethod}

In this section, we present a mobile framework design for infrared temperature measurement. To detect on mobile devices by a specialized thermal camera, a lightweight detection model is proposed and deployed below.

\subsection{Facial Landmark Detection}
Facial landmark detection is an important part of infrared temperature measurement.  With the help of facial landmarks, we can obtain body temperature from a more accurate area. 

Due to the influence of COVID-19, more and more people are used to wearing face masks. As depicted in Fig.~\ref{fig:markedImage}, The temperature of these areas covered by face masks or glasses is lower than other areas. Therefore, the temperature of a random area or the entire face area cannot be recognized as the human body temperature.

There are 5 most critical points on the face, including the left and right corners of the mouth, the center of the two eyes, and the nose. These points are the internal key points of the face. Since the forehead area is the least likely to be covered in the current scene, we prefer to use the temperature of this area as the human body temperature. The localization method of these 5 points will be shown in section~\ref{sec:MTCNNWithMaskDetection}.

\subsection{Image Alignment}
\begin{figure}[htbp]
	\centering
	\includegraphics[width=2.6in]{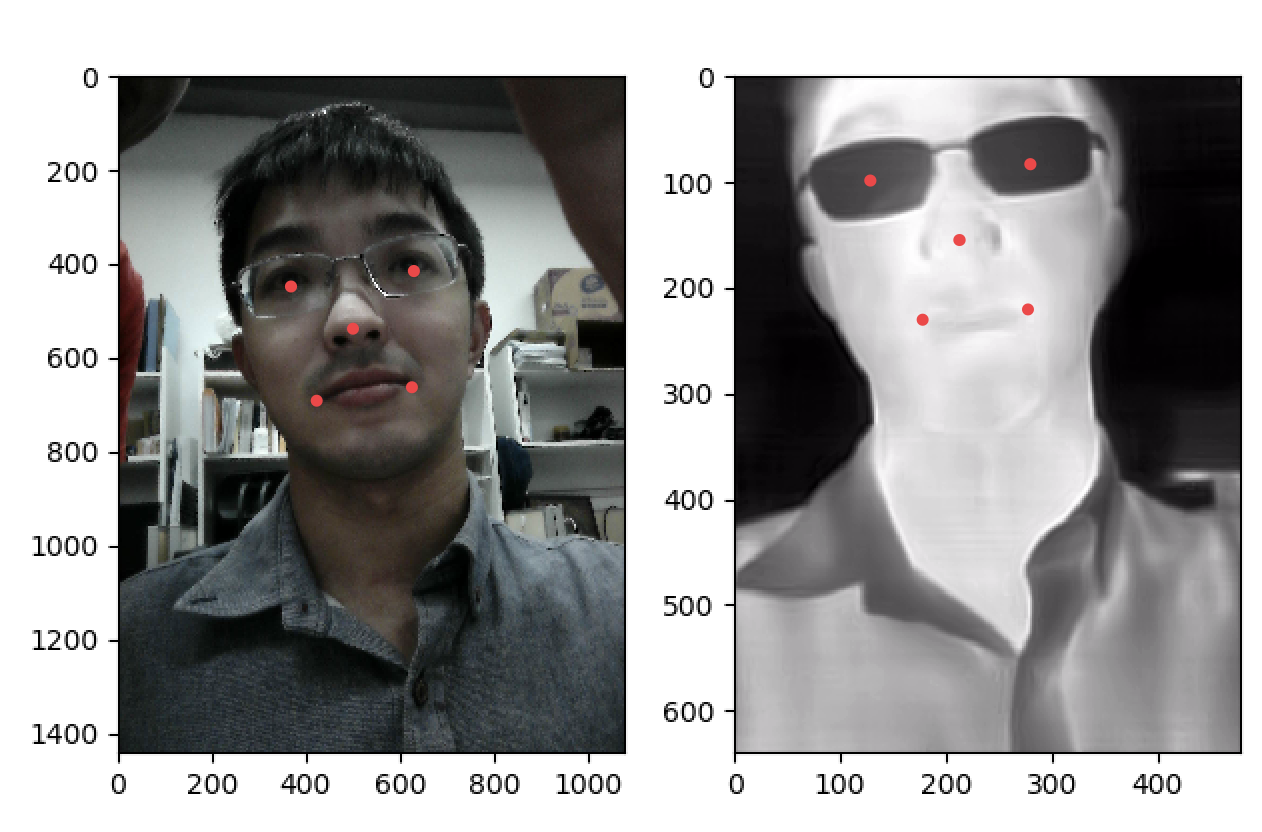}
	\caption{A pair of images with landmarks taken by FLIR ONE Pro.}
	\label{fig:markedImage}
\end{figure}

As shown in Fig.~\ref{fig:markedImage}, a pair of RGB and thermal images have different image resolutions. Besides, binocular disparity refers to the difference in image location of an object seen by the left and right eyes. Similarly, there is a disparity between a pair of images taken by a binocular camera because the two lenses cannot overlap in physical space. We cannot directly apply the landmark coordinate $(x, y)$ on the RGB image to the thermal image because of the resolution and disparity.

However, the spatial positions of the two lenses are fixed. That is, the straightness and parallelism of the RGB image will not change in the thermal image. Therefore, we can use affine transformation to align a pair of RGB and thermal images. An affine transformation includes scaling, translation, rotation, reflection, and shearing. It can be depicted as:

\begin{equation} 
	\label{equ:affine}
	\begin{bmatrix}
		x^\prime \\
		y^\prime \\
		1 
	\end{bmatrix}
	=
	A
	\begin{bmatrix}
		x \\
		y \\
		1
	\end{bmatrix}
	=
	\begin{bmatrix}
		a_1 & a_2 & t_x \\
		a_3 & a_4 & t_y \\
		0 & 0 & 1
	\end{bmatrix}
	\begin{bmatrix}
		x \\
		y \\
		1
	\end{bmatrix},
\end{equation}
where the coordinate $(x, y)$ represents the landmark on the RGB image and $(x^\prime, y^\prime)$ represents the landmark on the thermal image. An affine transformation matrix $A$ transforms $(x, y)$  to $(x^\prime, y^\prime)$. The translation of landmarks is controlled by parameters $t_x$ and $t_y$. Other transformations are controlled by parameter $a_{1-4}$. If these six unknown parameters are determined, the affine transformation matrix $A$ can be determined. We can use the matrix to locate the facial landmarks on the thermal image according to the facial landmarks detected on the RGB image. Afterwards, the temperature can be accurately measured.

We propose a loss calculation method to evaluate the accuracy of the affine transformation matrix. As shown below, Eq.~\eqref{equ:xLoss} represents the loss $L_{x}$ on the x-axis, Eq.~\eqref{equ:yLoss} represents the loss $L_{y}$ on the y-axis and Eq.~\eqref{equ:euclidLoss} represents the Euclidean distance loss $L_{euc}$. 

\begin{equation} 
	\label{equ:xLoss}
	L_{x} = \sum_i^n |x^\prime_i - x^{pred}_{i}|/(width_{t}\times n)
\end{equation}

\begin{equation} 
	\label{equ:yLoss}
	L_{y} = \sum_i^n|y^\prime_i - y^{pred}_{i}|/(height_{t}\times n)
\end{equation}

\begin{equation} 
	\label{equ:euclidLoss}
	L_{euc} = \sum_i^n\sqrt{(x^\prime_i-x^{pred}_{i})^2+(y^\prime_i-y^{pred}_{i})^2}/(diag_{t}\times n)
\end{equation}
where $n$ denotes the number of point pairs, ($x^\prime_i$, $y^\prime_i$) is the coordinate of a marked point, and ($x^{pred}_{i}$, $y^{pred}_{i}$) is the coordinate of the prediction result. Parameters $width_t$, $height_t$, and $diag_t$ are the width, height and diagonal length of the thermal image, respectively.

\begin{figure*}[htbp]
	\centering
	\includegraphics[scale=0.55]{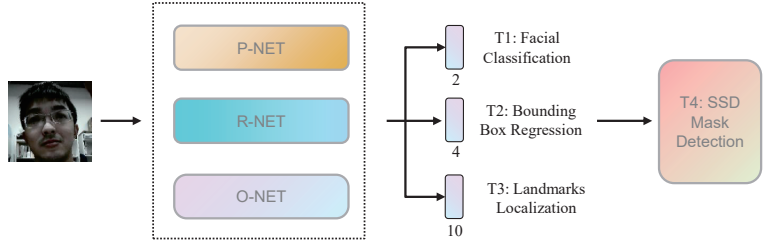}
	\caption{Overview of MTCNN with a mask detection branch.}
	\label{fig:mtcnnWithMask}
\end{figure*}

\section{MTCNN with a Mask Detection Branch}
\label{sec:MTCNNWithMaskDetection}
In this section, we propose a two-stage model based on multi-task cascaded convolutional networks (MTCNN)~\cite{MTCNN} and a single shot multibox detector (SSD)~\cite{SSD} detection model. The task of MTCNN is face detection and face alignment, while mask detection is the goal of SSD. Since both MTCNN and SSD are  lightweight architectures, they are suitable for running on mobile devices. With this model, we can find facial areas with a smaller coverage area to improve temperature measurement results.

\subsection{MTCNN Backbone}
MTCNN has a cascaded architecture, which has face detection and face alignment results. The output of MTCNN includes 3 tasks: face classification, bounding box regression, and facial landmark localization. The cascaded architecture contains three nets: P-Net, R-Net, and O-Net.  The layers of each net are shown in TABLE~\ref{tab:mtcnnArchitecture}.

\begin{table}[htbp]
	\caption{Layers of MTCNN proposed in \cite{MTCNN}}
	\begin{center}
		\begin{tabular}{|c|c|c|c|c|c|}
			\hline
			\textbf{Layer Number} & \textbf{P-Net} & \textbf{R-Net}& \textbf{O-Net} \\
			\hline
			Input Size & 12$\times$12$\times$3 & 24$\times$24$\times$3 & 48$\times$48$\times$3 \\
			\hline
			Layer 1 & Conv:3$\times$3 & Conv:3$\times$3 & Conv:3$\times$3  \\
			& MP:2$\times$2 & MP:3$\times$3 & MP:3$\times$3  \\
			\hline
			Layer 2 & Conv:3$\times$3 & Conv:3$\times$3 & Conv:3$\times$3  \\
			&  & MP:3$\times$3 & MP:3$\times$3  \\
			\hline
			Layer 3 & Conv:3$\times$3 & Conv:2$\times$2 & Conv:3$\times$3  \\
			&  &  & MP:2$\times$2  \\
			\hline
			Layer 4& - & FC128 & Conv:2$\times$2 \\
			\hline
			Layer 5& - & - & FC256 \\
			\hline
		\end{tabular}
		\label{tab:mtcnnArchitecture}
	\end{center}
\end{table}

MTCNN detects faces from coarse to fine. A cascaded architecture means that the output of the previous network is the input of the next network. The previous network gives a rough judgment firstly, and quickly deletes the areas that do not contain faces. The results are filtered by the next complex network to obtain accurate areas which contain faces.

The output of all three subnets is the same, including facial classification, bounding box regression, and facial landmark localization. As illustrated in Fig.~\ref{fig:mtcnnWithMask}, for each sample $x_i$, MTCNN has three tasks:

\subsubsection{Facial classification} 
This task contains 2 units as output, representing the probability of whether the image is a face. The loss function of facial classification $L_{i}^{\mathrm{cls}}$ is~\cite{MTCNN}: 
	
\begin{equation} 
	\label{equ:cls}
	L_{i}^{\mathrm{cls}}=-\left(y_{i}^{\mathrm{cls}} \log \left(p_{i}\right)+\left(1-y_{i}^{\mathrm{cls}}\right)\left(1-\log \left(p_{i}\right)\right)\right),
\end{equation}
where $p_i$ is the probability that sample $x_i$ is a face and $y_{i}^{\mathrm{cls}} \in \{0,1\}$ means the ground-truth label.
	
\subsubsection{Bounding box regression}  
This task contains 4 units as output, representing the top-left coordinate, width, and height of the bounding box. The loss function of bounding box regression $L_{i}^{\mathrm{box}}$ is~\cite{MTCNN}: 
	
\begin{equation} 
	\label{equ:bbox}
	L_{i}^{\mathrm{box}}=\left\|\hat{y}_{i}^{\mathrm{box}}-y_{i}^{\mathrm{box}}\right\|_{2}^{2}, 
\end{equation}
where $\hat{y}_{i}^{\mathrm{box}}$ is the prediction coordinate and ${y}_{i}^{\mathrm{box}}$ is the ground-truth coordinate. There are four parameters, including left, top, height, and width.
	
\subsubsection{Facial landmark localization} 
This task contains 10 units as output, representing the coordinates of 5 facial landmarks. The loss function of facial landmark localization $L_{i}^{\text {landmark}}$ is~\cite{MTCNN}: 

\begin{equation} 
	\label{equ:landmark}
	L_{i}^{\text {landmark}}=\left\|\hat{y}_{i}^{\text {landmark}}-y_{i}^{\text {landmark}}\right\|_{2}^{2},
\end{equation}
where $\hat{y}_{i}^{\mathrm{landmark}}$ is the prediction coordinate and ${y}_{i}^{\mathrm{landmark}}$ is the ground-truth coordinate. There are ten parameters, including the coordinates of the left and right corners of the mouth, the center of the two eyes, and the nose.

\subsection{Mask Detection Branch}
To avoid the spread of the virus, wearing masks in public areas is recommended. We add a mask detection branch after the MTCNN backbone. For making the model run in mobile devices, the detection branch uses an SSD architecture network to detect whether pedestrians wear masks on their faces.

SSD is a lightweight one-stage detection model, which can quickly and accurately detect masks. The location and classification layers are designed as TABLE~\ref{tab:ssd}.
\begin{table}[htbp]
	\caption{SSD Anchor Configuration}
	\begin{center}
		\begin{tabular}{|c|c|c|c|c|c|}
			\hline
			\textbf{Multibox Layers} & \textbf{Feature Map Size} & \textbf{Anchor Size}& \textbf{Aspect Ratio} \\
			\hline
			Layer 1 & 33$\times$33 & 0.04, 0.056 & 1, 0.62, 0.42  \\	
			\hline
			Layer 2 & 17$\times$17 & 0.08, 0.11 & 1, 0.62, 0.42  \\
			\hline
			Layer 3 & 9$\times$9 & 0.16, 0.22 & 1, 0.62, 0.42  \\
			\hline
			Layer 4& 5$\times$5 & 0.32, 0.45 & 1, 0.62, 0.42 \\
			\hline
			Layer 5& 3$\times$3 & 0.64, 0.72 & 1, 0.62, 0.42 \\
			\hline
		\end{tabular}
		\label{tab:ssd}
	\end{center}
\end{table}

In this branch, we add another new task mask detection, the loss function is similar to facial classification. We use a cross entropy loss function, which is often used in classification problems. The mask detection loss function $L_{i}^{\mathrm{m}}$ is:
\begin{equation} 
	\label{equ:m}
	L_{i}^{\mathrm{m}}=-\left(y_{i}^{\mathrm{m}} \log \left(p_{i}\right)+\left(1-y_{i}^{\mathrm{m}}\right)\left(1-\log \left(p_{i}\right)\right)\right),
\end{equation}
where $p_i$ is the probability that sample $x_i$ wearing a mask and $y_{i}^{\mathrm{m}} \in \{0,1\}$ means the ground-truth label.


\section{Experiments and Analysis}
\label{sec:ExperimentsAndAnalysis}
In this section, we first estimate the affine transformation matrix of the binocular thermal camera. The random sample consensus (RANSAC) estimation method, proposed in~\cite{RANSAC}, is applied to estimate the transformation matrix from our self-made image pairs. Then, the face alignment model with a mask branch is trained. Finally, we develop a TensorFlow Lite model that can be run on Android devices and detection results are uploaded to the cloud-edge-terminal collaborative system in real time. 

\subsection{Affine Transformation Matrix}
The RANSAC-based robust method is applied to calculate the affine transformation matrix of image pairs captured by the binocular camera. RANSAC is a robust estimation with a two-stage process: i) Classify data points as outliers or inliers. ii) Fit model to inliers while ignoring outliers. In this experiment, we only use a few datasets to calculate the affine transformation matrix and get a good result. 

We utilize RANSAC to fit the affine transformation by two sets of point matches: $source$ and $target$. The set $source$ is the coordinate set of points marked on the RGB image and the set $target$ is the coordinate set of points marked on the thermal image. Equation~\eqref{equ:affine} shows that affine transformation has 6 degrees of freedom, so a pair of images and 3 point matches are adequate to estimate the affine transformation matrix. To reduce manual error produced in the marking process, we marked more point matches.

\begin{figure}[htbp]
	\centering
	\includegraphics[width=2.6in]{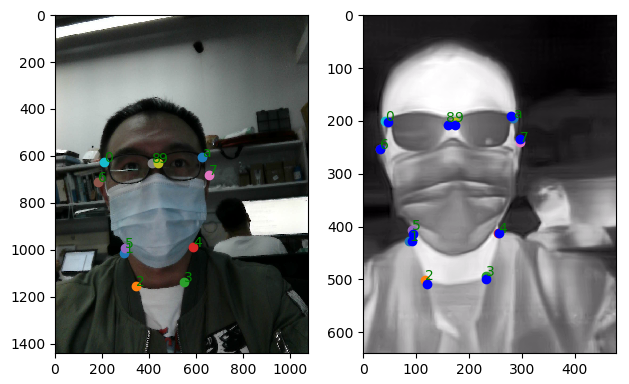}
	\caption{Transformation result (blue points on the thermal image are prediction result).}
	\label{fig:transformationResult}
\end{figure}

As shown in Fig.~\ref{fig:transformationResult}, we marked 20 pairs of RGB and thermal images for image alignment, and about 10 point matches will be marked on each pair of images. 10 pairs of RGB and thermal images are used for estimation and the other 10 pairs are for testing. 

The marked point matches of 10 pairs of RGB and thermal images, $source$ and $target$, are the input of the RANSAC method. And we get the affine transformation matrix $A$ of FLIR ONE Pro:

\begin{equation} 
	A
	=
	\begin{bmatrix}
		0.5584 & -0.0062 & -65.9722 \\
		-0.0014 & 0.5770 & -156.8899 \\
		0 & 0 & 1
	\end{bmatrix} \notag
\end{equation}

A detection result can be seen in Fig.~\ref{fig:transformationResult}, we apply the affine transformation matrix to the test dataset. On the left is an RGB image with marked points, and on the right is a thermal image with marked points and prediction results. As we can see, blue points cover the marked points on the thermal image, which means that the prediction result estimated by the affine transformation matrix is close to the correct result.

Matrix $A$ is applied to estimate the landmarks on the thermal images in our test dataset, and the transformation losses of the test thermal images are presented in TABLE~\ref{tab:transformationResult}. $L_x$, $L_y$ and $L_{euc}$ are all less than 1\%, which means the transformation result is close to the actual result.

\begin{table}[htbp]
	\caption{Transformation Loss}
	\begin{center}
		\begin{tabular}{|c|c|c|c|c|c|}
			\hline
			\textbf{Figure Number} & \textbf{$L_x$} & \textbf{$L_y$}& \textbf{$L_{euc}$} \\
			\hline
			Figure 1 & 4.9\textperthousand & 5.8\textperthousand & 6.2\textperthousand \\
			\hline
			Figure 2 & 4.4\textperthousand & 9.3\textperthousand & 8.5\textperthousand  \\
			\hline
			Figure 3 & 3.9\textperthousand & 3.7\textperthousand & 4.2\textperthousand  \\
			\hline
			Figure 4 & 1.8\textperthousand & 6.3\textperthousand & 5.2\textperthousand  \\
			\hline
			... & ... & ... & ... \\
			\hline
			\textbf{Average} & 3.9\textperthousand & 7.1\textperthousand & 6.7\textperthousand \\
			\hline
		\end{tabular}
		\label{tab:transformationResult}
	\end{center}
\end{table}

\subsection{Facial Landmarks and Temperature Measurement}

\setcounter{subsubsection}{0}\subsubsection{MTCNN Backbone  Training} Official WIDER FACE training set in~\cite{widerface} and LFW training set in~\cite{lfw} are used for training facial classification, bounding box regression, and facial landmark localization. WIDER FACE contains over 12000 training images and LFW contains 5590 training images.

After training, the loss of classification is 0.1144, the loss of bounding box is 0.05194 and the loss of landmarks is 0.01991. And in the WIDER FACE testing dataset, the accuracy of face detection is 85.1\%.

\subsubsection{Mask Detection Branch Training} In the mask detection branch training, 7959 training images with mask annotations are used. These training images are from the WIDER FACE dataset and added mask annotations. We use open-source mask datasets LFW, AgeDB-30 and CFP-FP from~\cite{maskDataset} as the test datasets. There are approximately 10,000 test images in each of the three test datasets. The accuracy, precision, and recall of the mask detection branch are calculated as follows: 


\begin{equation} 
	\label{equ:accuracy}	
	Accuracy = \frac{TP+TN}{TP+FP+FN+TN}
\end{equation}

\begin{equation} 
	\label{equ:precision}
	Precision = \frac{TP}{TP+FP}
\end{equation}

\begin{equation} 
	\label{equ:recall}
	Recall = \frac{TP}{TP+FN}
\end{equation}

\begin{figure}[htbp]
	\centering
	\includegraphics[width=2.6in]{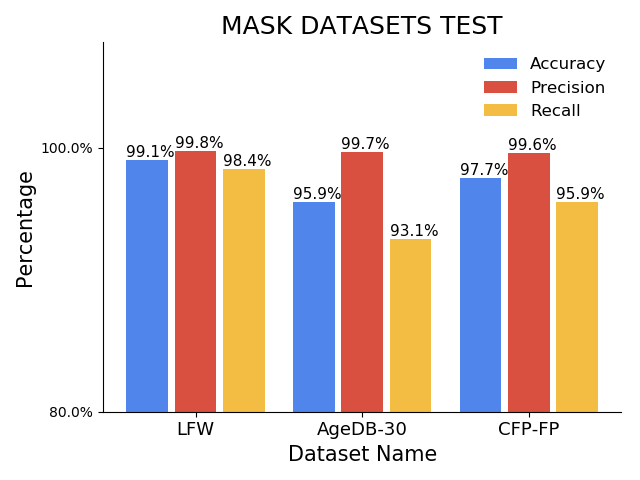}
	\caption{Test results on three mask datasets(LFW, AgeDB-30, and CFP-FP).}
	\label{fig:masktestresult}
\end{figure}

In Eqs.~\eqref{equ:accuracy},~\eqref{equ:precision}, and~\eqref{equ:recall}, $TP$ represents the number of true positive samples, $TN$ represents the number of true negative samples, $FP$ represents the number of false positive samples, and $FN$ represents the number of false negative samples. The test result is presented in Fig.~\ref{fig:masktestresult}, the accuracy of the model on the three test datasets is higher than 95\%. 


\subsubsection{Temperature Measurement} Figure~\ref{fig:transformationResult} shows that if you wear a face mask and glasses, most of the face will be covered. 
Thus, we cannot measure the temperature of these covered areas. 
For selecting the forehead which is least covered, we need the bounding box and facial landmark results to locate the best temperature measurement area.
We take the coordinates of 4 points, including the top-left and top-right corners of the bounding box, the left eye and the right eye.  We connect the top-left corner and the right eye, and connect the top-right corner and the left eye. The intersection of the two straight lines is the center of the temperature measurement area.

\subsection{System Deployment}
\begin{figure}[ht]
	\centering
	\includegraphics[width=2.4in]{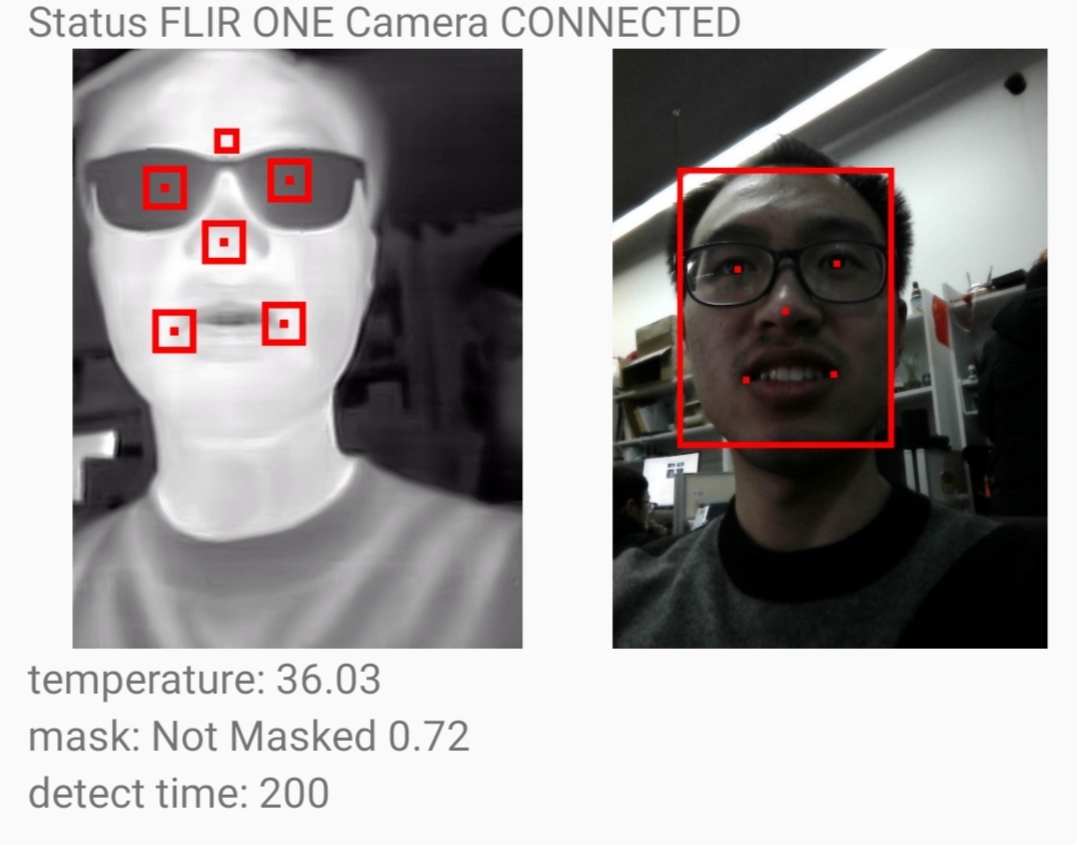}
	\caption{A real-time detection experiment.}
	\label{fig:androidapp}
\end{figure}

In order to easily detect the body temperature of pedestrians, we need a model that can be run on a mobile platform. Thus, we use TensorFlow Lite to convert the detection model to a mobile detection model. Our experiment platform is Honor V30 with Kirin 980 SoC. A real-time detection experiment result is presented in Fig.~\ref{fig:androidapp}.

The final size of the entire model is 6.1M. An Android application is developed to apply this mobile model by Android Studio, and the average detection speed is  257ms. At a distance of 1m, the error of indoor temperature measurement is about 3\%. Once a face is detected, the location of the mobile device and the temperature are uploaded to the cloud.

\section{Conclusions}
\label{sec:Conclusion} 
An accurate mobile body temperature measurement system is proposed in this paper. The cloud-edge-terminal system ensures the timeliness of data transmission and the stability of the temperature measurement system. 
In addition, we calculate the affine transformation matrix of the image pairs taken by the binocular camera. The transformation result, which has only a 6.7\textperthousand\ Euclidean error, is close to the actual result. Then, a lightweight face alignment model with a mask detection branch is proposed to realize detection on mobile devices. The average detection speed of our Android model is 257ms, and the error of indoor temperature measurement at a distance of 1m is about 3\% , which means that our model can realize real-time and accurate measurement in public areas. Finally, the information can be uploaded to the cloud in real time.

In the future, we plan to prune the network and get a faster and smaller mobile model which will make more contributions to COVID-19 prevention and control.

\bibliographystyle{IEEEtran}
\bibliographystyle{unsrt}
\bibliography{Reference}

\end{document}